\pdfoutput=1
\documentclass[conference]{IEEEtran}
\IEEEoverridecommandlockouts
\usepackage{cite}
\usepackage{amsmath,amssymb,amsfonts}
\usepackage{graphicx}
\usepackage{textcomp}
\usepackage{xcolor}
\usepackage{multirow}
\usepackage{url}
\usepackage{booktabs}
\usepackage{microtype}
\def\BibTeX{{\rm B\kern-.05em{\sc i\kern-.025em b}\kern-.08em
    T\kern-.1667em\lower.7ex\hbox{E}\kern-.125emX}}
\begin{document}

\title{INSIGHT: Indoor Scene Intelligence from Geometric-Semantic Hierarchy Transfer for Public~Safety\\
\thanks{Public Safety Communications Research Division, National Institute of Standards and Technology.}
}

\author{%
\IEEEauthorblockN{Alexander Nikitas Dimopoulos}
\IEEEauthorblockA{\textit{Innovation Lab} \\
\textit{PSCR, NIST} \\
Boulder, CO, USA \\
alexander.dimopoulos@nist.gov \\
ORCID: 0009-0001-8742-3940}
\and
\IEEEauthorblockN{Joseph Grasso}
\IEEEauthorblockA{\textit{Location Based Services} \\
\textit{PSCR, NIST} \\
Boulder, CO, USA \\
joseph.grasso@nist.gov \\
ORCID: 0009-0000-6075-2958}
\and
\IEEEauthorblockN{John Beltz}
\IEEEauthorblockA{\textit{Network Operations Group} \\
\textit{PSCR, NIST} \\
Gaithersburg, MD, USA \\
john.beltz@nist.gov \\
ORCID: 0009-0005-5932-5760}
}

\maketitle

\begin{abstract}
Indoor environments lack the spatial intelligence that GPS provides outdoors: a first responder arriving at an unfamiliar building typically has no machine-readable map of its safety equipment. Prior work on 3D semantic segmentation for public safety identified two barriers. Labeled indoor training data is scarce, and methods that operate directly on point clouds recognize small safety-critical fixtures poorly. This paper presents INSIGHT, a pipeline that transfers 2D image understanding into 3D metric space using registered color-and-depth (RGB-D) imagery, with no manual annotation of the target building. Two interchangeable vision stacks share a common 3D back end: one built on the Segment Anything Model~3 (SAM3) foundation model for text-prompted segmentation, and one built from traditional computer-vision (CV) components---open-set detection, visual question answering (VQA), and optical character recognition (OCR)---whose intermediate outputs are independently inspectable. Evaluated on all seven subareas of Stanford 2D-3D-S (70{,}496 images), INSIGHT produces labeled point clouds compatible with the Pointcept training schema and scene graphs aligned with the ISO~19164 indoor feature model, reducing the field payload by roughly $10^{4}{\times}$ relative to the source geometry; graphs filtered by responder role transmit in ${<}15$\,s at 1\,Mbps over FirstNet Band~14, the dedicated United States public-safety broadband network. We report per-point labeling accuracy on the 7 classes shared with Stanford ground truth, detection volume for 15 safety-critical classes that public 3D benchmarks annotate sparsely or not at all, together with code-referenced deployable estimates, and complementarity between the two stacks. The results demonstrate that 2D-to-3D semantic transfer addresses the labeled-data bottleneck, while scene graphs deliver building intelligence compact enough for field use.
\end{abstract}


\section{Introduction}

Outdoor environments are saturated with spatial intelligence: GPS, satellite imagery, routable street networks. Indoor environments lack comparable data. A first responder arriving at an unfamiliar structure typically relies on static floor plans, if any exist, that rarely encode the operational features needed most, such as where standpipe cabinets are mounted, which panel controls utility shutoff, or whether an Automated External Defibrillator (AED) is available in the north corridor. The National Fire Protection Association's (NFPA) 1660 standard (Ch.~17)~\cite{NFPA1660} mandates pre-incident planning for the structures a department may be called to, and NFPA~950~\cite{NFPA950} requires that the resulting data carry spatial attributes in interoperable form. In practice both are bottlenecked by manual data production.

Our prior analysis of NIST Point Cloud City~\cite{Dimopoulos2025,NIST_PCC} identified two barriers. First, labeled indoor training data is scarce and covers almost none of the safety-critical infrastructure responders need. Second, native 3D models such as KPConv~\cite{Thomas2019} score near-zero Intersection over Union (IoU) on small safety features (fire extinguishers, exit signs, electrical panels) that lack distinctive geometry at typical scan resolutions. Foundation models trained on internet-scale imagery address the recognition problem in 2D; when registered with metric depth, 2D understanding transfers into 3D space without manual annotation.

This paper presents INSIGHT (Indoor Scene Intelligence from Geometric-Semantic Hierarchy Transfer), which lifts 2D image understanding into 3D building intelligence using registered depth data. INSIGHT exports Pointcept-schema-compatible~\cite{Wu2024} labeled point clouds and ISO~19164~\cite{ISO_19164_2024} / IndoorGML-aligned~\cite{IndoorGML} scene graphs sized for public-safety broadband. The design is dual-track: SAM3~\cite{Carion2025} for breadth and zero-shot recognition of the safety classes that no standard training set labels, and a traditional CV stack (YOLOE~\cite{Wang2025} $\rightarrow$ BLIP visual question answering (VQA)~\cite{Li2022} $\rightarrow$ PaddleOCR optical character recognition (OCR)~\cite{Du2020}) for modular, stage-inspectable detection---aligning with the NIST Artificial Intelligence Risk Management Framework (AI RMF)~\cite{NIST_AIRMF} principles.

\textbf{Contributions.}
\begin{enumerate}
\item A dual-pipeline architecture that transfers 2D semantics into 3D with zero target-domain annotation, producing both Pointcept-schema training data and ISO-19164-aligned scene graphs from either vision stack. The modular CV stack serves as a stage-inspectable cross-check on SAM3, and disagreement between the stacks acts as an operational review signal, consistent with NIST AI RMF~\cite{NIST_AIRMF} auditability principles.
\item A systematic characterization of where a foundation-model stack and a classical CV stack diverge, on a 23-class public-safety taxonomy spanning 70,496 images and seven subareas across three buildings. The stacks agree on general classes and diverge sharply on safety-critical ones, which they detect at rates differing by up to 50$\times$ and on which they overlap by 0--20\%, below 7\% for ten of the 14 safety classes compared.
\item The first detection-volume baseline for 15 safety-critical classes---AEDs, fire alarm panels, standpipes, utility shutoffs---ten of which no public 3D benchmark annotates at all, and the remainder only coarsely (ScanNet200 carries \texttt{fire extinguisher}, \texttt{fire alarm}, \texttt{stairs}, \texttt{rail}).
\item A standards-aligned scene-graph format that reduces the field payload by 20{,}000--27{,}000$\times$ relative to the source geometry and fits sub-megabit public-safety links: role-filtered payloads deliver in ${<}15$\,s at 1\,Mbps over FirstNet Band~14.
\end{enumerate}

\section{Background}\label{sec:background}

\subsection{Public Safety Context}\label{sec:context}

Standards define both the demand and the format for this work. NFPA~1660's~\cite{NFPA1660} pre-incident planning mandate creates the demand; NFPA~950~\cite{NFPA950} points the resulting data toward interoperable exchange formats such as IFC~\cite{IFC2024}, ISO~19164~\cite{ISO_19164_2024}, and IndoorGML~\cite{IndoorGML}. INSIGHT automates data generation at exactly that intersection.

The delivery constraint is FirstNet Band~14~\cite{FirstNetBand14}, the dedicated United States public-safety broadband band. NIST TN~1552~\cite{NIST_TN1552} measures in-building attenuation at 750\,MHz in four large structures; at the path losses it reports, sub-megabit effective throughput is the operating assumption we adopt. At 1\,Mbps, a 30-second fireground decision window---an operational planning assumption based on NFPA~1660 size-up procedures~\cite{NFPA1660}---yields a payload budget of roughly 3.75\,MB.

The NIST AI RMF~\cite{NIST_AIRMF} requires that AI in high-consequence settings be explainable and accountable, motivating INSIGHT's inspectable CV cross-check (\S\ref{sec:cv_stack}) and the inter-stack disagreement signal (\S\ref{sec:complementarity}).

\subsection{Related Work}\label{sec:related}

Large indoor benchmarks---S3DIS~\cite{Armeni2017}, ScanNet~\cite{Dai2017}, and SemanticKITTI~\cite{Behley2021}---advanced 3D segmentation but annotate almost none of the classes public-safety responders require. Native 3D methods (KPConv~\cite{Thomas2019}, Point Transformer~\cite{Zhao2021}, Stratified Transformer~\cite{Lai2022}) therefore remain gated by this data gap, which our prior work~\cite{Dimopoulos2025} quantified on small safety features. SAM~3~\cite{Carion2025} extends the SAM~\cite{Kirillov2023} / SAM~2~\cite{Ravi2024} line with text-prompted concept segmentation, critical for classes absent from any training set. 2D-to-3D transfer systems (OpenMask3D~\cite{Takmaz2024}, OVIR-3D~\cite{Lu2023}, Open3DIS~\cite{Nguyen2024}, Segment3D~\cite{Huang2024}, OpenScene~\cite{Peng2023}, ConceptGraphs~\cite{Gu2024}) project 2D features into 3D but target benchmark evaluation and dense reconstruction; INSIGHT instead targets operational deployment: a fixed safety taxonomy, ISO-19164-aligned output, a $10^{4}{\times}$ payload reduction, and dual outputs for training and field use.

Operational indoor systems---DHS/Mappedin Response~\cite{DHS_Mappedin}, NEVERLOST~\cite{Sangenis2025}, and others~\cite{TRACLabs_FRIA,Fire360}---target \emph{live} incident response. INSIGHT occupies the complementary \emph{pre-incident} slot, generating the scene-graph layer live systems can consume~\cite{Gamboa2024}. Armeni \emph{et al.}~\cite{Armeni2020_SG} and Wald \emph{et al.}~\cite{Wald2020} established 3D scene-graph vocabularies for residential and office reasoning, but no prior system combines automated vision-based detection, an NFPA-compatible safety taxonomy, and ISO-19164 output for first-responder deployment.

\section{Methodology}\label{sec:method}

\subsection{System Overview}\label{sec:overview}

\begin{figure*}[htbp]
  \centering
  \includegraphics[width=\textwidth]{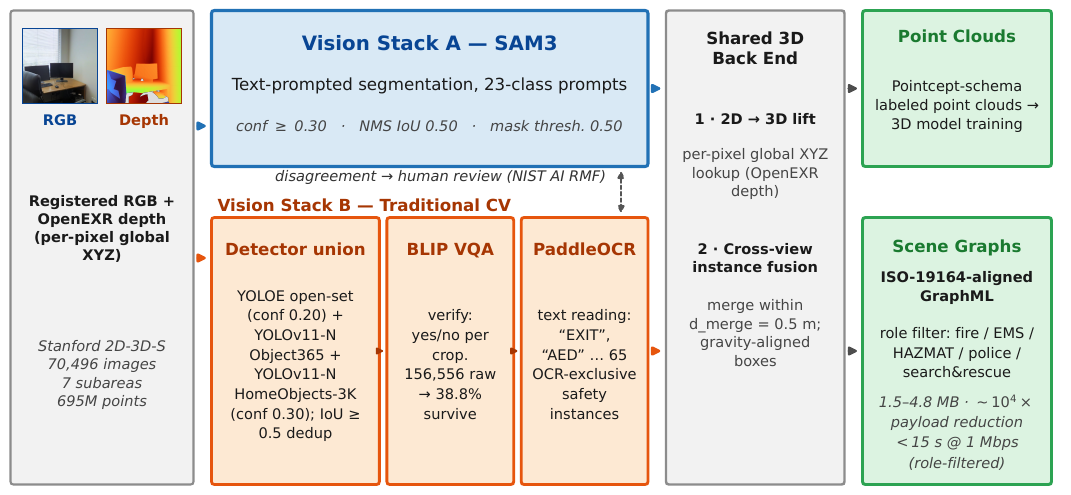}
  \caption{INSIGHT pipeline. Registered RGB and OpenEXR depth (70{,}496 images, 7 subareas) flow through two interchangeable vision stacks: Stack~A, SAM3 text-prompted segmentation over the 23-class taxonomy; and Stack~B, a modular CV chain in which three pre-trained/open-set detectors run in union topology, BLIP VQA verifies candidate crops, and PaddleOCR reads text-dependent fixtures. A shared 3D back end looks up per-pixel global XYZ from depth, fuses instances across views ($d_\text{merge}{=}0.5$\,m), and exports (a) Pointcept-schema-compatible labeled point clouds for model training and (b) ISO-19164-aligned, role-filterable scene graphs (${\sim}10^{4}{\times}$ smaller payload) for field delivery. Stack~B's inspectable intermediate outputs cross-check Stack~A; inter-stack disagreement flags human review, per NIST AI RMF~\cite{NIST_AIRMF}.}
  \label{fig:architecture}
\end{figure*}

Each pixel in the Stanford 2D-3D-S~\cite{Armeni2017} OpenEXR depth map encodes the world-space $(x,y,z)$ of the corresponding RGB pixel, bridging 2D segmentation to 3D placement. The pipeline (Fig.~\ref{fig:architecture}) processes images through one of two interchangeable vision stacks, projects detections into 3D, fuses instances across viewpoints, and exports both labeled point clouds and scene graphs.

Off-the-shelf datasets label almost none of the safety infrastructure that NFPA~1660 (Ch.~18) requires. In the configuration evaluated here, safety-class coverage therefore comes from open-vocabulary prompting---SAM3 text prompts in Stack~A, YOLOE class prompts and OCR in Stack~B---rather than from any safety-supervised detector. Fine-tuning a dedicated safety detector and integrating it into Stack~B is future work (\S\ref{sec:discussion}).

\subsection{Vision Stack A: SAM3 Foundation Model}\label{sec:sam3}

SAM3 receives one text prompt per target class and produces pixel-level masks with confidence scores. Detections below 0.30 are discarded; NMS with IoU 0.50 removes overlaps; mask threshold 0.50 hardens soft predictions. Most prompts restate the class name (``fire extinguisher,'' ``electrical panel,'' ``exit sign''); three are broader than the class they populate (\texttt{fire\_alarm\_pull}~$\leftarrow$~``fire alarm,'' \texttt{gas\_shutoff}~$\leftarrow$~``gas valve,'' \texttt{water\_shutoff}~$\leftarrow$~``water valve''), so those three classes absorb related fixtures---ceiling alarm appliances in particular---and their counts should be read as upper bounds.

\subsection{Vision Stack B: Traditional CV Pipeline}\label{sec:cv_stack}

The CV stack is deliberately \emph{modular}: an open-set detector, a unified verification stage, and a text reader, each replaceable and independently inspectable---a stage-level cross-check on SAM3 (\S\ref{sec:complementarity}).

\textbf{Detection.} All three detectors run on every image in a \emph{union} topology; their outputs are merged with IoU-based deduplication ($\text{IoU}\geq0.5$) after verification. The primary detector is YOLOE~\cite{Wang2025}, an open-set YOLOv11~\cite{Jocher2024}-Large variant that accepts class prompts at inference (confidence threshold 0.20; 56.9\% of kept detections). Two YOLOv11-Nano detectors complement it at threshold 0.30: one pre-trained on Object365 (6.7\%) and one on HomeObjects-3K (36.4\%), which supplies furniture, doors and windows but no safety classes. Every safety-class detection in Stack~B therefore originates with YOLOE or with OCR. Per-view detection statistics in this subsection---model shares and the verification funnel below---are measured on Area~6 (9{,}812 images), the subarea whose per-detection log was retained; all instance counts reported elsewhere in the paper cover the full seven subareas. All three detectors run at the Ultralytics default $640{\times}640$ input on Stanford's native $1{,}080{\times}1{,}080$ frames, with no tiled or multi-scale inference---the configuration a department deploying an off-the-shelf detector would run. Tiling (e.g., SAHI) or a larger input would improve small-fixture recall at higher compute cost.

\textbf{Unified verification.} BLIP VQA~\cite{Li2022} validates YOLOE candidates via yes/no questions per crop; Object365 and HomeObjects detections are deduplicated against VQA-verified YOLOE outputs before entering the final pool. Of 156,556 raw detections, 38.8\% (60,798) survive; VQA rejects 15,422 (9.9\%), and class filtering, duplicate removal and the confidence thresholds account for the remainder. Two-thirds of those rejections carry furniture-mapped labels (lamp, TV, cabinet) rather than safety classes; the remainder include 322 elevators, 136 fire alarm pulls and 105 fire hose cabinets. Which rejections were correct is unverified---no per-crop review was performed---so the stage's cost to safety classes is a count, not a measured error rate. This trade-off motivates the fusion argument in \S\ref{sec:complementarity}.

\textbf{Text reading.} PaddleOCR~\cite{Du2020} extracts text-dependent fixtures (``EXIT,'' ``AED,'' ``FIRE DEPT VALVE''); 65 of its 154 safety detections fall where no visual model placed a candidate (\S\ref{sec:safety}).

\subsection{3D Projection, Instance Fusion, and Taxonomy}\label{sec:projection}

Both stacks share the 3D back end. For each detection, world coordinates are extracted from the OpenEXR depth map; a global instance registry merges detections whose 3D centroids fall within $d_\text{merge}=0.5$\,m. Each instance receives a gravity-aligned bounding box and confidence equal to the max across observations. The 23-class taxonomy (Table~\ref{tab:taxonomy}) extends S3DIS's 13 structural classes with egress, fire suppression, fire alarm, utility control, and medical categories; structural classes follow IFC~\cite{IFC2024} naming conventions (\S\ref{sec:scenegraph}).

\begin{table}[htbp]
\caption{Public Safety Class Taxonomy (23 classes, by operational function).}
\label{tab:taxonomy}
\centering
\scriptsize
\setlength{\tabcolsep}{3pt}
\renewcommand{\arraystretch}{1.0}
\begin{tabular}{|c|l|c|}
\hline
\textbf{Category} & \textbf{Classes} & \textbf{Count} \\
\hline
Egress \& Access & door, window, stairs, elevator, ramp, exit\_sign, railing & 7 \\
\hline
Fire Suppression & fire\_extinguisher, standpipe, fire\_hose\_cabinet, sprinkler & 4 \\
\hline
Fire Alarm & fire\_alarm\_panel, fire\_alarm\_pull & 2 \\
\hline
Utility Control & electrical\_panel, gas\_shutoff, water\_shutoff & 3 \\
\hline
Medical & aed & 1 \\
\hline
Structural & wall, floor, ceiling, column & 4 \\
\hline
Obstacles & furniture, clutter\footnotemark & 2 \\
\hline
\end{tabular}
\end{table}
\footnotetext{Clutter is defined for taxonomy completeness but not actively targeted.}

\subsection{Role-Based Scene Graph Generation}\label{sec:scenegraph}

Scene graphs encode building structure as an IndoorGML-aligned hierarchy (Building~$\rightarrow$~Floor~$\rightarrow$~Surfaces~$\rightarrow$~Instances), with each node storing semantic class, ISO class name, 3D oriented box, confidence, and responder priority, exported as GraphML.

\textbf{ISO~19164 scope.} We use \emph{aligned} deliberately: INSIGHT implements part of the conceptual core of the ISO~19164 indoor feature model~\cite{ISO_19164_2024}, not a certified encoding. Implemented are a two-level containment hierarchy (Building~$\rightarrow$~Floor) consistent with the cell-space concept but without room-level subdivision; structural surfaces as cell-space boundary features; IFC-style~\cite{IFC2024} entity names for structural classes, though safety classes use pipeline-specific names; and safety attributes (responder priority, discipline relevance) as extra node fields rather than a formal subclass schema. Not implemented are the connectivity/routing dual, the multi-layered space model, and GML encoding (we serialize to GraphML).

\textbf{Role and priority assignment.} Discipline roles and priority levels are assigned by a fixed, author-defined lookup table in the pipeline source~\cite{InsightRepo2026}, covering the 22 classes actively carried into the detection registry (\texttt{clutter}, listed in Table~\ref{tab:taxonomy} for taxonomy completeness, is not detected and has no registry entry). Each class maps to the disciplines that act on it---firefighting, hazmat, police/tactical, EMS, search-and-rescue---and to a priority tier informed by the pre-incident data elements NFPA~1660 (Ch.~18)~\cite{NFPA1660} calls for. The mapping is a configuration file, not a learned component, so agencies can re-map to local doctrine without modifying the pipeline.

A query-time role filter emits discipline-specific views: a firefighter-filtered SAM3 graph retains 12,639 of 30,086 nodes (58\% reduction); EMS retains 5,589 (81\%).

\section{Results}\label{sec:results}

\subsection{Dataset and Metrics}\label{sec:dataset}

We evaluate on all seven subareas (1--6, with 5 split into 5a/5b per community convention) of Stanford 2D-3D-S~\cite{Armeni2017}: 70,496 RGB+OpenEXR images, 695\,M source points, ${>}6{,}000$\,m$^2$. The \emph{overlapping set} (7 classes) admits per-point accuracy; the \emph{novel set} (15 safety-critical classes) lacks GT and is assessed by proxy signals (\S\ref{sec:safety}). Implementation: PyTorch~${\geq}2.0$ mixed-precision on an NVIDIA A100; SAM3 weights from \texttt{facebookresearch/sam3} at float16.

\textbf{Novel-class validation.} The 15 safety-critical classes lack per-point ground truth, so their precision is \emph{unverified} here; we report proxy signals that bound plausibility instead (\S\ref{sec:safety}).

\subsection{Per-Point Labeling Accuracy}\label{sec:accuracy}

\begin{table*}[htbp]
\caption{Per-point semantic accuracy (\%) on 7 overlapping classes against Stanford 2D-3D-S ground truth, over all 7 subareas. Best per class in bold. Column accuracy is near zero for both stacks because the area-wide wall surface claims the points Stanford labels column or beam (\S\ref{sec:accuracy}). $^\ddagger$GT instance counts from Stanford pointcloud.mat; Area~5 is annotated once in Stanford's 3D data and counted once here. $^\dagger$Spatial coverage: \% of pipeline surface points within 0.1\,m of any GT point (Area~3 diagnostic).}
\label{tab:iou_results}
\centering
\scriptsize
\setlength{\tabcolsep}{4pt}
\renewcommand{\arraystretch}{1.05}
\begin{tabular}{|c|c|c|c|c|c|c|c|c|}
\hline
\textbf{Pipeline} & \textbf{Ceiling} & \textbf{Floor} & \textbf{Wall} & \textbf{Column} & \textbf{Window} & \textbf{Door} & \textbf{Furniture} & \textbf{Overall} \\
\hline
GT Instances$^\ddagger$ & 385 & 284 & 1{,}546 & 413 & 168 & 543 & 2{,}457 & 5{,}796 \\
\hline
SAM3         & 93.9 & \textbf{94.0} & \textbf{95.9} & \textbf{1.4} & 64.1 & \textbf{68.4} & 84.7 & \textbf{80.6} \\
\hline
CV Pipeline  & \textbf{95.2} & 86.1 & 46.0 & 0.7 & \textbf{83.4} & 63.3 & \textbf{98.5} & 77.8 \\
\hline
Spatial Coverage$^\dagger$ & 99.8 & 99.7 & 99.9 & --- & --- & --- & --- & --- \\
\hline
\end{tabular}
\end{table*}

\textbf{Metric definition.} Per-point accuracy is computed over the 7 classes shared between the pipeline taxonomy and Stanford 2D-3D-S after class mapping (table$+$chair$+$sofa$+$bookcase $\rightarrow$ furniture; beam $\rightarrow$ column; six classes map directly). For each pipeline point whose semantic label maps to one of these 7 classes, the nearest Stanford GT point in 3D supplies the reference label. A point is \emph{correct} if the two mapped labels agree. Per-class values in Table~\ref{tab:iou_results} are computed over the points whose nearest GT point carries that class. Points carrying pipeline-only labels (the 15 novel safety classes) are excluded; Stanford-only classes (board, clutter) have no pipeline equivalent and are excluded from the denominator. Each stored instance and surface contributes points in proportion to its true size, so the area-wide structural surfaces---two-thirds of the labeled cloud---are weighted as they actually occur. The reported overall accuracy is the point-count-weighted mean of per-area accuracies.

\begin{figure}[htbp]
  \centering
  \includegraphics[width=\columnwidth]{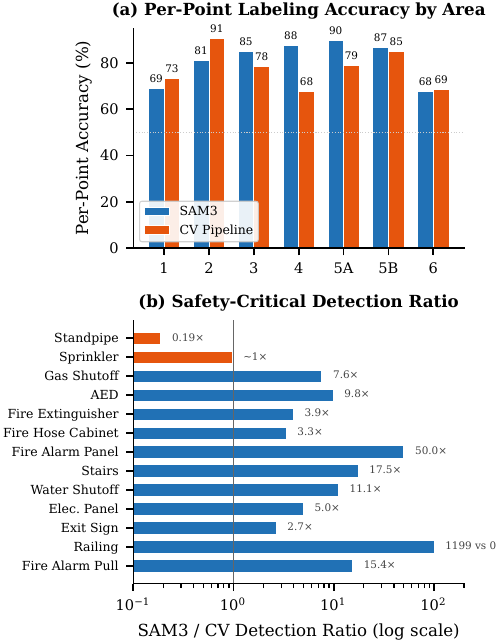}
  \caption{(a) Per-area per-point accuracy; SAM3 leads in four of seven subareas. (b) SAM3/CV safety detection ratio (log scale; the railing bar, undefined because CV detects none, is truncated at the axis limit): alarm and utility classes favor SAM3 by 3--50$\times$; standpipe (0.19$\times$) is a dual-pipeline failure (\S\ref{sec:safety}).}
  \label{fig:accuracy_safety}
\end{figure}

SAM3 achieves 80.6\% area-weighted per-point accuracy; CV, 77.8\% (Table~\ref{tab:iou_results}, Fig.~\ref{fig:accuracy_safety}a). SAM3 leads on wall, floor and door and in four of the seven subareas; CV leads on ceiling, window and furniture. Excluding furniture, the gap widens to 80.3\% against 68.0\%, and SAM3's advantage is concentrated in wall (95.9\% vs.\ 46.0\%). Our prior KPConv analysis on NIST Point Cloud City~\cite{Dimopoulos2025} (IoU, a stricter metric on a different dataset, so not numerically comparable) converges on the same qualitative ordering: large structural surfaces are the tractable case, small safety features the hard one.

\textbf{Column caveat.} Column is the one class where both stacks fail outright (1.4\% and 0.7\%). Representing each structural surface as a single area-wide instance means the wall surface spatially subsumes most columns and pilasters, so points Stanford labels column or beam are returned as wall. This is a representational choice in the scene-graph export rather than a detection failure---the stacks do register column instances (527 and 332, Table~\ref{tab:fragmentation})---but it makes the per-point metric uninformative for that class, and room-level column placement should not be inferred from the labeled cloud.

\subsection{Safety-Critical Detection: The AED Gap and the 50$\times$ Alarm Ratio}\label{sec:safety}

\begin{table}[htbp]
\caption{Novel safety detections (fused 3D instances), all areas. No GT exists; counts reflect detection volume, not verified instances. SAM3 covers all 7 areas except sprinkler (6/7) and standpipe (3/7). CV counts include PaddleOCR contributions (154 total). Thirteen of the 15 novel classes appear: ramp yields 1 SAM3 instance and 0 CV, and elevator (SAM3 85, CV 607, 0.14$\times$) is reported in Fig.~\ref{fig:complementarity}; including elevator, the total ratio is 4.4$\times$. The railing ratio is undefined (CV detects none).}
\label{tab:novel_detections}
\centering
\scriptsize
\setlength{\tabcolsep}{3pt}
\renewcommand{\arraystretch}{1.0}
\begin{tabular}{|l|r|r|r|}
\hline
\textbf{Class} & \textbf{SAM3} & \textbf{CV} & \textbf{Ratio} \\
\hline
fire\_alarm\_pull   & 1,568 &   102 & 15.4$\times$ \\
\hline
railing             & 1,199 &     0 & --- \\
\hline
exit\_sign          &   926 &   348 &  2.7$\times$ \\
\hline
electrical\_panel   &   704 &   142 &  5.0$\times$ \\
\hline
water\_shutoff      &   566 &    51 & 11.1$\times$ \\
\hline
stairs              &   437 &    25 & 17.5$\times$ \\
\hline
fire\_alarm\_panel  &   350 &     7 & \textbf{50.0$\times$} \\
\hline
fire\_hose\_cabinet &   291 &    87 &  3.3$\times$ \\
\hline
fire\_extinguisher  &   280 &    71 &  3.9$\times$ \\
\hline
aed                 &   167 &    17 &  9.8$\times$ \\
\hline
gas\_shutoff        &   114 &    15 &  7.6$\times$ \\
\hline
sprinkler           &    26 &    27 & $\sim$1$\times$ \\
\hline
standpipe           &     3 &    16 &  0.19$\times$ \\
\hline
\textbf{Total}      & \textbf{6,631} & \textbf{908} & \textbf{7.3$\times$} \\
\hline
\end{tabular}
\end{table}

Both pipelines detect safety-critical objects absent from Stanford's annotations (Table~\ref{tab:novel_detections})---classes on which KPConv scored near-zero IoU in our prior analysis~\cite{Dimopoulos2025} (0.005 on exit signs, 0.000 on AEDs). Across the 13 novel safety classes in Table~\ref{tab:novel_detections}, SAM3 detects 6,631 instances versus CV's 908 (7.3$\times$). The gap is widest for alarm and utility classes: fire alarm panels 350 vs.\ 7 (\textbf{50$\times$}), fire alarm pulls 1,568 vs.\ 102 (15.4$\times$), AEDs 167 vs.\ 17 (9.8$\times$). Standpipe is the only class in Table~\ref{tab:novel_detections} where CV materially leads (16 vs.\ 3; sprinkler is at parity), but counts this low across seven subareas represent a dual-pipeline failure rather than a CV advantage---standpipes' visual appearance is likely underrepresented in foundation-model pretraining. Elevator, reported in Fig.~\ref{fig:complementarity}, is the one novel class CV dominates (607 vs.\ 85).

The AED gap is illustrative: all 17 CV-detected AEDs come from PaddleOCR reading ``AED'' on mounted cabinets. Across five safety classes, OCR contributes 154 high-confidence instances (mean 0.903), of which 65 are exclusive---not duplicating visual detections but addressing a modality gap.

VQA removes the bulk of off-taxonomy proposals, but its cost falls on the classes the pipeline exists to find (\S\ref{sec:cv_stack}), motivating the fusion analysis below.

Two proxy signals bound novel-class precision without per-instance annotation. (1)~\emph{Inter-pipeline agreement}: at 1\,m centroid match radius (\S\ref{sec:complementarity}), safety classes exhibit 0--20\% overlap (below 7\% for ten of 14) between stacks that share no model weights or detection logic, so co-detections constitute independent corroboration. (2)~\emph{Residual over-count against code}: 570 SAM3 instances survive confidence gating ($\tau{=}0.70$), against a code-minimum expectation of 126 across the seven subareas (Table~\ref{tab:prior_filter}), so gating alone leaves a 4.5$\times$ residual. The cited provisions prescribe minimum fixture provision rather than maxima, so the top-$K$ step is a presentation cap, not evidence of correctness. Definitive precision requires stratified sampling against source RGB crops, scoped for deployment pilots, where NFPA~1660 (Ch.~17)~\cite{NFPA1660} already mandates domain-expert review.

\subsection{Pipeline Complementarity}\label{sec:complementarity}

\begin{figure*}[htbp]
  \centering
  \includegraphics[width=0.72\textwidth]{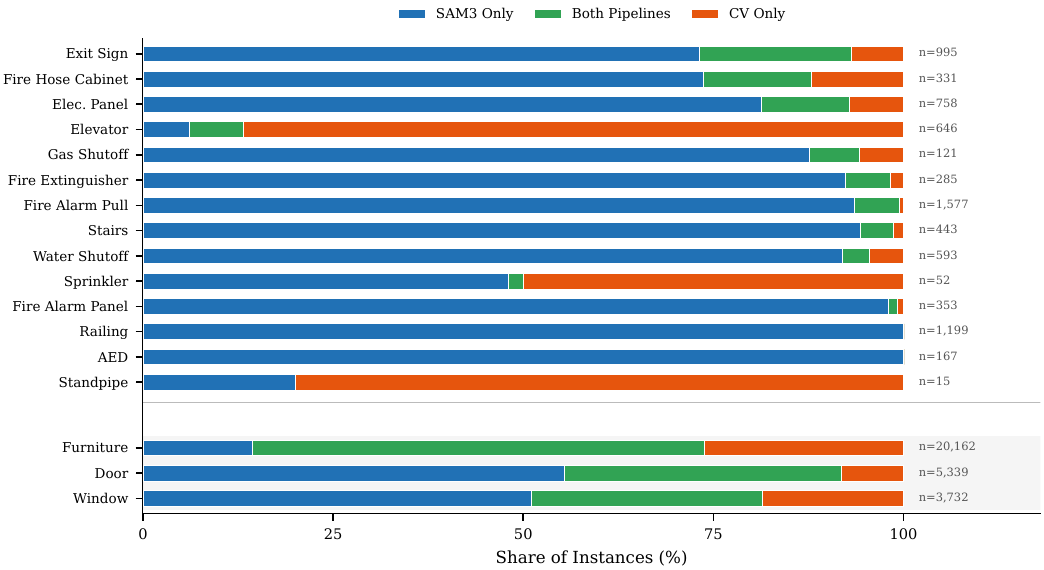}
  \caption{Detection complementarity at 1\,m 3D-centroid match radius, shown as share of instances per class. General classes exhibit substantial overlap (furniture 59\%), whereas safety-critical classes show low agreement (0--20\%; below 7\% for ten of the 14). Structural surfaces are excluded (single-instance-per-area representation), as is ramp (1 SAM3 instance, 0 CV).}
  \label{fig:complementarity}
\end{figure*}

At 1\,m centroid match radius (Fig.~\ref{fig:complementarity}), 41.4\% of instances are detected by both pipelines, 38.4\% by SAM3 only, 20.1\% by CV only, yielding 37,645 unique instances, 25\% more than SAM3 alone (structural surfaces and ramps excluded). Safety classes show low overlap (exit signs 19.9\% at the high end; alarm pulls 5.9\%, alarm panels 1.1\%, AEDs 0\%), so fusion captures most of the sum of both: the stacks fail differently and succeed on different viewpoints.

\subsection{Confidence-Threshold Stability}\label{sec:threshold}

\begin{figure}[htbp]
  \centering
  \includegraphics[width=\columnwidth]{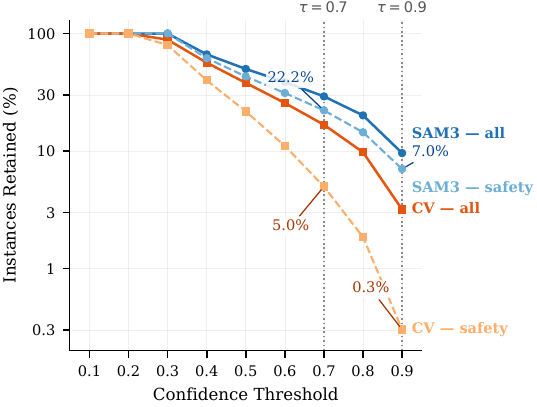}
  \caption{Instance retention vs.\ confidence threshold. Solid: all classes; dashed: the safety-relevant subset, which here also includes door, window and stairs ($n{=}13{,}372$ SAM3, $4{,}958$ CV; OCR instances excluded)---a wider set than the 7 capped classes of Table~\ref{tab:prior_filter}. At $\tau{=}0.7$ the CV pipeline retains 5.0\% of this subset (SAM3: 22.2\%); at $\tau{=}0.9$, 0.3\% (SAM3: 7.0\%).}
  \label{fig:threshold}
\end{figure}

Fig.~\ref{fig:threshold} quantifies the consequence of the high-confidence gating liability-conscious departments may require: at threshold 0.7 the CV pipeline retains only 5.0\% of the safety-relevant subset (SAM3: 22.2\%); at 0.9, CV retains 0.3\% (SAM3: 7.0\%). Restricted to the 7 classes capped in Table~\ref{tab:prior_filter}, the same gate retains 13.3\% for SAM3 and 2.4\% for CV. These CV figures, like Table~\ref{tab:prior_filter}, exclude the OCR instances of Table~\ref{tab:novel_detections}, 136 of whose 154 exceed 0.7; OCR is the one CV component whose confidence survives conservative gating. SAM3's higher and more stable confidence on safety classes is essential for any deployment applying conservative thresholds.

\subsection{Multi-View Fragmentation and Plausibility Filtering}\label{sec:fragmentation}

Per-view detection without appearance-based re-identification produces systematic overcounting. Table~\ref{tab:fragmentation} reports the \emph{count ratio} (pipeline instances divided by GT instances) for the 4 discrete overlapping classes (structural surfaces are excluded because both pipelines compress them to one instance per area by design).

\begin{table}[htbp]
\caption{Count ratio (pipeline detections\,/\,reference count) at $d_{\text{merge}}{=}0.5$\,m. Reference is Stanford GT for overlapping classes---Area~5 is annotated once in Stanford's 3D data and counted once here---and the code-minimum reference $K$ for novel classes. The ratio conflates fragmentation, false positives and misses, so a near-unity value need not indicate agreement (see column, below).}
\label{tab:fragmentation}
\centering
\scriptsize
\setlength{\tabcolsep}{3pt}
\renewcommand{\arraystretch}{1.0}
\begin{tabular}{|l|r|r|r|r|r|}
\hline
\textbf{Class} & \textbf{Ref.} & \textbf{SAM3} & \textbf{S/Ref} & \textbf{CV} & \textbf{C/Ref} \\
\hline
\multicolumn{6}{|l|}{\textit{Overlapping (ref = GT instances):}} \\
\hline
door       & 543   & 4{,}899 & 9.0$\times$ & 2{,}379 & 4.4$\times$ \\
window     & 168   & 3{,}041 & 18.1$\times$ & 1{,}825 & 10.9$\times$ \\
column     & 413   & 527     & 1.3$\times$ & 332     & 0.8$\times$ \\
furniture  & 2{,}457 & 14{,}881 & 6.1$\times$ & 17{,}259 & 7.0$\times$ \\
\hline
\multicolumn{6}{|l|}{\textit{Novel safety (ref = plausibility cap $K$):}} \\
\hline
fire\_alarm\_pull   & 21  & 1{,}568 & 74.7$\times$ & 102 & 4.9$\times$ \\
fire\_alarm\_panel  & 7   & 350     & 50.0$\times$ & 7   & 1.0$\times$ \\
exit\_sign          & 35  & 926     & 26.5$\times$ & 348 & 9.9$\times$ \\
aed                 & 7   & 167     & 23.9$\times$ & 17  & 2.4$\times$ \\
\hline
\end{tabular}
\end{table}

Overlapping-class ratios range from 0.8$\times$ (columns, CV) to 18.1$\times$ (windows, SAM3); novel-class ratios reach 50--75$\times$ for SAM3 alarm classes, confirming that the plausibility filter below is essential. The column ratio is the clearest warning against reading these values as agreement: at $1.3\times$ (SAM3) it reflects frequent misses offset by frequent false positives rather than one detection per fixture. The merge distance $d_{\text{merge}}{=}0.5$\,m is a conservative default; a systematic ablation is deferred to future work.

Raw counts in Table~\ref{tab:novel_detections} exceed plausible fixture counts (167 AEDs, 350 alarm panels) because fragmentation and low-confidence false positives compound. We apply a two-stage post-hoc filter (Table~\ref{tab:prior_filter}): (1)~a confidence gate at $\tau=0.70$ (\S\ref{sec:threshold}), then (2)~a cardinality cap retaining the top-$K$ detections per subarea-class pair, where $K$ is the fixture count the cited codes would lead one to expect for a building of this size (Table~\ref{tab:prior_filter}). Because those provisions prescribe minimum provision rather than maxima, $K$ is a reference count applied as a presentation cap, not a bound the data is shown to satisfy. SAM3's 4{,}286 raw safety detections reduce to 125 (97\% reduction) without fabricating any instances. Raw counts measure \emph{detection capability}; filtered counts, the \emph{responder-facing view}.

The cap is an interim mitigation: transparent in principle (each retained instance derives from identifiable source views, though the current export does not persist the instance-to-view link), conservative (it removes but never creates instances), and grounded in code rather than tuned on the data. It does not fix the underlying failure---the absence of cross-view identity, since detections of one fixture merge only when their 3D centroids fall within $d_\text{merge}$, so localization error beyond 0.5\,m spawns duplicates. Appearance-based re-identification is the principled remedy (\S\ref{sec:discussion}).

\begin{table}[htbp]
\caption{Plausibility filter on safety-critical classes. $\tau=0.70$ confidence gate followed by a top-$K$ cap, with $K$ a code-minimum reference count per NFPA~10~\cite{NFPA10} (extinguishers), NFPA~72~\cite{NFPA72} (pulls, panels), NFPA~14~\cite{NFPA14} (hose cabinets), and AHA~\cite{AHA_AED} (AEDs), summed across the 7 subareas; filtered count $= \min(\text{surviving}, K)$. Since these provisions set minima, equality with $K$ reflects the cap binding rather than agreement with code. CV columns count visual-detector instances only; the 154 OCR instances of Table~\ref{tab:novel_detections} (136 above $\tau$) are excluded.}
\label{tab:prior_filter}
\centering
\scriptsize
\setlength{\tabcolsep}{3pt}
\renewcommand{\arraystretch}{1.0}
\begin{tabular}{|l|r|r|r|r|r|}
\hline
\textbf{Class} & \textbf{SAM3 raw} & \textbf{$\geq 0.70$} & \textbf{$K$} & \textbf{SAM3 filt.} & \textbf{CV filt.} \\
\hline
aed                 & 167   & 6   & 7  & 6  & 0 \\
\hline
fire\_alarm\_panel  & 350   & 18  & 7  & 7  & 0 \\
\hline
fire\_alarm\_pull   & 1{,}568 & 226 & 21 & 21 & 10 \\
\hline
fire\_extinguisher  & 280   & 34  & 21 & 21 & 1 \\
\hline
fire\_hose\_cabinet & 291   & 90  & 14 & 14 & 1 \\
\hline
exit\_sign          & 926   & 165 & 35 & 35 & 2 \\
\hline
electrical\_panel   & 704   & 31  & 21 & 21 & 1 \\
\hline
\textbf{Subtotal}   & \textbf{4{,}286} & \textbf{570} & \textbf{126} & \textbf{125} & \textbf{15} \\
\hline
\end{tabular}
\end{table}

\subsection{Scene-Graph Payload Reduction and Delivery}\label{sec:compression}

\begin{table}[htbp]
\caption{Per-area scene-graph payload reduction (SAM3). CV achieves 9,600--16,000$\times$ with 1.3--3.9\,MB graphs. Ratios computed from unrounded byte counts.}
\label{tab:scene_graph}
\centering
\scriptsize
\setlength{\tabcolsep}{3pt}
\renewcommand{\arraystretch}{1.0}
\begin{tabular}{|c|r|r|r|r|}
\hline
\textbf{Area} & \textbf{Geo DB (GB)} & \textbf{Graph (MB)} & \textbf{Ratio} & \textbf{Nodes} \\
\hline
Area 1  &  86.1 & 4.2 & 20,498$\times$ & 4,857 \\
Area 2  & 126.4 & 4.6 & 27,340$\times$ & 5,345 \\
Area 3  &  31.3 & 1.5 & 21,399$\times$ & 1,694 \\
Area 4  & 121.6 & 4.7 & 26,054$\times$ & 5,394 \\
Area 5a &  55.4 & 2.8 & 20,094$\times$ & 3,181 \\
Area 5b &  96.6 & 4.8 & 20,206$\times$ & 5,505 \\
Area 6  &  83.3 & 3.6 & 23,475$\times$ & 4,110 \\
\hline
\end{tabular}
\end{table}

\begin{table}[htbp]
\caption{FirstNet-tier transmission time for a representative mid-size area (Area 1, 86.1\,GB geometry, 4.2\,MB full graph). Role-filtered sizes are measured on the generated filtered graphs. MB${=}10^{6}$\,bytes.}
\label{tab:firstnet}
\centering
\scriptsize
\setlength{\tabcolsep}{4pt}
\renewcommand{\arraystretch}{1.0}
\begin{tabular}{|l|r|r|r|r|}
\hline
\textbf{Payload} & \textbf{Size} & \textbf{@1\,Mbps} & \textbf{@5\,Mbps} & \textbf{@25\,Mbps} \\
\hline
Raw geometry DB          & 86.1\,GB & 191\,h & 38.2\,h & 7.65\,h \\
Full SAM3 scene graph    &  4.2\,MB & 33.6\,s & 6.7\,s & 1.3\,s \\
Firefighter-filtered    &  1.80\,MB & \textbf{14.4\,s} & 2.9\,s & 0.6\,s \\
EMS-filtered            &  0.75\,MB & \textbf{6.0\,s} & 1.2\,s & 0.2\,s \\
\hline
\end{tabular}
\end{table}

SAM3 reduces the payload 20,000--27,000$\times$, from geometry databases (31--127\,GB) to scene graphs (1.5--4.8\,MB, Table~\ref{tab:scene_graph}); CV achieves 9,600--16,000$\times$. The graph discards geometry rather than encoding it compactly, so this is a payload reduction, not a recoverable compression. Against the 30-second fireground window at 1\,Mbps (a 3.75\,MB budget, \S\ref{sec:context}), raw geometry misses by more than four orders of magnitude, the full SAM3 graph marginally exceeds it, and role-filtered graphs land well below (Table~\ref{tab:firstnet}).

\section{Discussion and Conclusion}\label{sec:discussion}

\textbf{Findings.} INSIGHT produces usable labeled point clouds with zero target-domain annotation, closing the labeled-data gap identified in our prior Point Cloud City analysis~\cite{Dimopoulos2025}. The two stacks are strongly complementary (safety-class overlap 0--20\%, \S\ref{sec:complementarity}). SAM3 provides broad safety coverage and stable high-confidence retention; the CV pipeline excels at furniture and elevators and provides a stage-inspectable second opinion. PaddleOCR contributes the CV stack's \emph{only} AED coverage (SAM3 detects AEDs visually); text is a distinct modality that the CV stack's visual detectors did not substitute for.

The dual pipeline is the architecture, not a comparison: CV's visual safety detection collapses above 0.7, leaving SAM3 and OCR alone, and this behavior should inform deployment-threshold policy under the NIST AI RMF~\cite{NIST_AIRMF}.

\textbf{Limitations.} Two limitations dominate. First, the 15 novel safety classes have no per-point ground truth, so their true precision is unverified: raw counts (Table~\ref{tab:novel_detections}) and filtered counts (Table~\ref{tab:prior_filter}) are non-GT-validated estimates of detection volume and deployable inventory, the proxy signals (\S\ref{sec:safety}) bound plausibility without establishing correctness, and the verification protocol defined there is still to be executed in deployment pilots. Three classes are populated by prompts broader than the class itself (\S\ref{sec:sam3}); \texttt{fire\_alarm\_pull} is the most affected, and its count should be read as an upper bound on alarm devices of all kinds rather than on manual pull stations. Second, multi-view fragmentation: both pipelines produce 0.8--18$\times$ the GT instance count on discrete classes (Table~\ref{tab:fragmentation}), the plausibility filter (\S\ref{sec:fragmentation}) treats the symptom rather than the cause, and a $d_{\text{merge}}$ ablation is still needed; appearance-based re-identification is the principled fix. Beyond these, spot-check inspection suggests room-level positional accuracy sufficient for awareness but not waypoint navigation (an observation, not a systematic measurement). Column accuracy is near zero in both stacks under per-point evaluation (\S\ref{sec:accuracy}), as the area-wide wall surface claims the points Stanford labels column or beam. Standpipe, ramp, and sprinkler detection is near-zero in both pipelines. Stack~B carries no safety-supervised detector (\S\ref{sec:overview}), so its safety coverage rests entirely on open-vocabulary prompting and OCR; a fine-tuned safety detector is the obvious next increment. Evaluation covers only Stanford 2D-3D-S; other building types are future work via NIST Point Cloud City~\cite{NIST_PCC}.

Several extensions would broaden the impact of automated point-cloud annotation for public safety:
\begin{enumerate}
\item \emph{Appearance-based re-identification} to collapse multi-view fragmentation (Table~\ref{tab:fragmentation}), the dominant error source: extract an appearance embedding per detection crop, associate detections across views by embedding similarity jointly with 3D proximity, and replace the fixed $d_\text{merge}$ merge with the learned association.
\item \emph{3D model training on INSIGHT pseudo-labels}, closing the annotation loop.
\item \emph{Ensemble fusion} treating inter-pipeline disagreement as a calibrated confidence signal.
\item \emph{IndoorGML connectivity edges} for routing---walls and floors define traversable-space boundaries (\S\ref{sec:accuracy}), and furniture defines obstacles.
\item \emph{Domain generalization} beyond public safety---from \emph{collect, annotate, train} to \emph{scan, detect, transfer, train}---to facilities management, indoor robotics, and BIM maintenance.
\end{enumerate}

INSIGHT translates modern computer vision into operational building intelligence: a 20{,}000--27{,}000$\times$ payload reduction, a 23-class NFPA-aligned taxonomy, and a sub-megabit delivery envelope around a dual-pipeline architecture. By automating what NFPA~1660 currently requires departments to produce by hand, INSIGHT shifts pre-incident planning from a data-production bottleneck to a deployable workflow. The same transfer pattern can let responders enter unfamiliar structures with machine-readable knowledge of egress, suppression, and medical resources, without the annotation burden that has stalled indoor spatial intelligence.

\section*{Acknowledgment}

Certain commercial equipment, instruments, software, or materials are identified in this paper to specify the experimental procedure adequately. Such identification does not imply recommendation or endorsement by the National Institute of Standards and Technology, nor does it imply that the materials or equipment identified are necessarily the best available for the purpose.

Claude Code~\cite{ClaudeCode2025}, an AI-assisted coding tool by Anthropic, was used in the development of the INSIGHT pipeline source code, analysis scripts, and the visualization scripts that generate Figs.~\ref{fig:architecture}--\ref{fig:threshold}.


\end{document}